\documentclass{bmvc2k}

\usepackage{hyperref}
\usepackage{url}
\usepackage{amsmath}
\usepackage{algorithm}
\usepackage{algpseudocode}
\usepackage{pifont}
\usepackage{color}
\usepackage{graphicx}
\usepackage{wrapfig}
\usepackage{caption}
\usepackage{subcaption}

\makeatletter
\def\BState{\State\hskip-\ALG@thistlm}
\makeatother

\usepackage{mathtools}
\DeclareMathOperator*{\argmin}{argmin}

\title{Subspace Alignment Based Domain Adaptation for RCNN Detector}

\addauthor{Anant Raj}{anantraj@iitk.ac.in}{1}
\addauthor{Vinay P. Namboodiri}{vinaypn@iitk.ac.in}{2}
\addauthor{Tinne Tuytelaars}{tinne.tuytelaars@esat.kuleuven.be}{3}

\addinstitution{
Department of Electrical Engineering,\\
 IIT Kanpur,\\
 Kanpur, India.
}
\addinstitution{
 Department of Computer Science and Engineering,\\
 IIT Kanpur,\\
 Kanpur, India.
}
\addinstitution{
 ESAT, PSI-VISICS\\
 KU Leuven,\\
 Heverlee, Belgium
}

\runninghead{Raj, Namboodiri, Tuytelaars}{Adapting RCNN Detector}

\begin{document}
\maketitle
\vspace{-0.5cm}
\begin{abstract}
\vspace{-0.1cm}
In this paper, we propose subspace alignment based domain adaptation of the 
state of the art RCNN based object detector \cite{ES10}. The aim is to
be able to achieve high quality object detection in novel, real world
target scenarios without requiring labels from the target domain. While,
unsupervised domain adaptation has been studied in the case of object
classification, for object detection it has been relatively unexplored. In
subspace based domain adaptation for objects, we need access to source and
target subspaces for the bounding box features. The absence of supervision
(labels and bounding boxes are absent) makes the task challenging. In
this paper, we show that we can still adapt subspaces that are localized
to the object by obtaining detections from the RCNN detector trained
on source and applied on target. Then we form localized subspaces from
the detections and show that subspace alignment based adaptation between these subspaces yields improved object detection. This evaluation is done by considering challenging real world datasets of PASCAL VOC as source and validation set of Microsoft COCO dataset as target for various categories.
\end{abstract}

\vspace{-0.7cm}
\section{Introduction}
\label{sec:intro}
\vspace{-0.2cm}
It has been an underlying assumption behind most of the machine learning algorithms that training and test instances should be sampled from a similar distribution. But this assumption is often violated in a real world scenario, i.e. there is high probability that train and test instances can arise from different distributions. This problem is well known as the domain shift problem in research community. The problem of domain shift is visible in various fields including language processing, speech processing as well as in computer vision tasks. Various factors can cause this problem in computer vision. For example, if somebody has trained the object classifiers on images being taken from a high quality DSLR camera and the test instances are taken from images being taken from a VGA camera then the performance of the classifiers is not supposed to be good at all. Apart from the difference in resolution, difference in view points, clutter and background can also cause the problem of domain shift. Indeed, the problem is particularly pertinent to the computer vision community due to our reliance on `standard' challenging datasets. Each dataset has its own bias and the results of one dataset do not easily transfer to other datasets as has been shown by Torralba and Efros in an important work \cite{ES18}. Due to these factors, domain adaptation task is becoming of higher importance in computer vision. However most of the work in this field revolves around adapting a classifier for the task of object recognition or classification and not much effort has been put to adapt an object detector. 

Through this paper, a contribution we make is to analyse the object detection performance between two challenging standard object detection datasets viz, Pascal VOC and Microsoft CoCo using the state of the art RCNN object detection technique. While, one would assume that the use of convolutional neural networks that have been trained with all the examples from the Imagenet dataset would result in the detector working well across datasets, we show that such is not the case. To adapt the object detection in the unsupervised setting is challenging. If we had observations from the other dataset, then we could use fine-tuning to adapt the convolutional neural network itself. Without having access to supervision, we consider a recent domain adaptation technique based on subspace alignment to adapt the feature subspaces between source and target subspaces for localized object detection bounding boxes. We further analyse our method by considering the principal angles between the subspaces. Our evaluation demonstrates that it is possible to obtain localized subspace adaptation for object detection and that this adaptation results in improved performance for off-the-shelf improved object detection.

Rest of the paper is organized as follows. In the next section we discuss the related work, and in section \ref{bacgrnd} the main concepts of subspace learning and detection are developed. In section \ref{our_appr} the proposed method has been discussed in detail. Experimental evaluation is presented in section \ref{expts} and it includes the performance evaluation and detailed analysis. We finally conclude in section \ref{concl}.


\vspace{-0.5cm}

\section{Related Work}
\label{rel_work}
\vspace{-3mm}
The task of visual domain adaptation for object classification has been studied in unsupervised and semi-supervised settings. In this section we briefly survey only the domain adaptation techniques that are related to the present work. A recent extended report \cite{SURVEY} by Gopalan {\it et al.} surveys domain adaptation techniques for visual recognition. 

Subspace based methods are commonly used for learning new feature representations that are domain-invariant. They thus enable the transfer of classifiers from a source to a specific target  domain. A common approach behind these subspace based methods is to first determine separate subspaces for source and target data. The data is then projected onto a set of intermediate sampled subspaces along the geodesic path between source and target subspace with the aim of making the feature point domain invariant \cite{ES1}. This approach has been further continued in terms of geodesic flow kernel \cite{ES2}, source and target subspace alignment \cite{ES3} and manifold alignment based approach \cite{ES4}. 

Different from subspace based techniques, in \cite{ES14}, it is shown that the image representation learned by convolutional neural networks on Imagenet dataset can be transferred for other tasks with relatively smaller dataset by using fine tuning the deep network. However, these require annotations for the target dataset. In very recent work a more sophisticated technique has been proposed by Zhang {\it et al} \cite{ES13} where a deep transfer network is learned that learns a shared feature subspace that matches conditional distributions. However, this is not applicable for detecting objects.


The above mentioned works are applicable for object classification. The problem of domain adaptation for object detection has been studied to a lesser extent. 
One such work applicable to object detection has been the adaptation of Deformable Part Models for object detection \cite{ES9}. 
The adaptation of objects through transfer component analysis \cite{ES17} has been proposed by Mirrashed {\it et al.} \cite{ES5} specifically for the case of vehicles. In recent work there has been a method proposed to adapt a fine tuned convolutional neural network for detection by considering it as a domain adaptation technique \cite{ES15}. This technique has shown interesting  performance for large scale detection. However, they heavily rely on the presence of a large number of pre-trained fine-tuned detectors (200 categories). We do not make such assumptions for our method.

Finally, our work is also motivated by the idea of hierarchical and iterative domain adaptation. In \cite{ES7}, it has been proposed to adapt the hierarchy of features to exploit the visual information. 
 In \cite{ES11}, Anant {\it et al.} propose an iterative hierarchical subspace based domain adaptation method to exploit the availability of additional structure in the label space i.e.\ hierarchy. However, these techniques are not applicable for detection.

\vspace{-5mm}

\section{Background}
\label{bacgrnd}
\vspace{-3mm}
The proposed approach builds upon the previously proposed subspace alignment based method \cite{ES3} for visual domain adaptation to adapt the RCNN detector \cite{ES10}. 
\vspace{-3mm}
\subsection{Subspace Alignment}
\label{SA}
\vspace{-0.3cm}
Subspace alignment based domain adaptation method consists of learning a transformation matrix $M$
that maps the source subspace to the target one \cite{ES1}. Suppose, we have labelled source data S and unlabelled target data T. We normalize the data vectors and take separate PCA of the source data vectors and target data vectors. The $d$ eigenvectors for each domain are selected corresponding to the
$d$ largest eigenvalues. We consider these eigenvectors as bases for source and target subspaces separately. They are denoted by $X_s$ for source subspace and $X_t$ for target subspace. We use a transformation matrix $M$ to align the source subspace $X_s$ to target subspace $X_t$. The mathematical formulation to this problem is given by 
\begin{eqnarray}
\vspace{-0.4cm}
F(M) = \|X_SM-X_T\|_F^2 \qquad  M^* = \argmin_M (F(M)).
\vspace{-0.4cm}
\label{equ1}
\end{eqnarray}
 $X_s$ and $X_t$ are matrices containing the $d$ most important eigenvectors for source and target respectively and $\|.\|_F$ is the \textit{Frobenius norm}. The solution of {eqn.} \ref{equ1} is $M^* = X_S'X_T$ and hence for the target aligned source coordinate system we get $X_a = X_SX_S'X_T$.  Once we get target aligned source co-ordinate system, we project our source data and train the classifier in this frame. While testing, target data is projected on the target subspace and classifier score is calculated.

\subsection{RCNN-detector}
\label{RCNN}
Convolutional neural nets (CNN) and other deep learning based approaches have improved the object classification accuracy by a large margin. RCNN \cite{ES10} uses the CNN framework and bridges the gap between object classification and object detection task. The idea of this work is to see how well the result of convolutional neural network on ImageNet task generalizes for the task of object detection on PASCAL dataset. RCNN consists of three modules. The first module generates selective search windows \cite{ES16} in an image which is category independent. Second module extracts mid level convolutional neural network features for each proposed region which has been trained earlier on ImageNet dataset. In the third module, SVM classifier is trained by considering all those windows whose overlap with the ground truth bounding box are less then a threshold $\lambda$ as negative examples. Hard negative examples are mined from these negative examples during the training. In testing phase, again 2000 selective search windows are generated per image in fast mode. Each proposal is warped and propagated forward through pre-trained  CNN to compute features.  Then,  for  each  class, the learned SVM class specific classifier is applied to those extracted features and a score is obtained corresponding to each proposal. Once we get the scores, we decide a threshold and the regions with scores greater than the decided threshold are our possible candidates for a particular object category. In the last step greedy non maximum suppression is applied to obtain desired, accurate and specific bounding box for that object category. 

\vspace{-0.6cm}
\section{Subspace Alignment for Adapting RCNN}
\label{our_appr}
\vspace{-0.2cm}

In this section we describe our approach to adapt the class specific RCNN -detector. On the basis of background provided in the previous section, we use subspace alignment based domain adaptation over the initial RCNN-detector.  Instead of using single subspace for the full source and target data, we postulate that using class-specific different subspaces for different classes to adapt from source to target domain improves the object detection accuracy.

\begin{algorithm}
\caption{Subspace Alignment based Domain Adaptation for RCNN Detector}\label{alg:euclid}
\begin{algorithmic}[1] 
\Procedure{SA based RCNN Adaptation}{Source Data S,Target Data T}
\For{each image $j$ \Pisymbol{psy}{206} $Source$ and $Target$ $Image$ }
\State $Windows(j)$ $\gets$ $ComputeSelectiveSearchWindows(j)$
\State $feat(j)$ $\gets$ $ComputeCaffeFeat(Windows(j))$
\EndFor
\State $InitRCNNdetector$ $\gets$ $TrainRCNNonSource(SourceData)$

\For{each class $i$ \Pisymbol{psy}{206} $Object$ $Class$ }
\State $PosSrc(i)$ = () and $PosTgt(i)$ = ()
\For{each image $j$ \Pisymbol{psy}{206} $Source$ and $Target$ $Image$}
\State $ol(j)$ = $ComputOverlap(gTBbox(j,i),Windows(i))$ \Comment For source images
\State $PosSrc(i)$ = Stack($PosSrc(i)$,$feat(i)(ol(j) \geq \gamma)$ \Comment For source images
\State $score(i,j)$ = $runInitRCNNdetector(image(j))$   \Comment For target images
\State $PosTgt(i)$ = Stack($PosTgt(i)$,$feat(i)(score(i,j) \geq \sigma)$ \Comment For target images
\EndFor

\State $X_{source}(i)$ $\gets$ $PCA(PosSrc(i))$
\State $X_{target}(i)$ $\gets$ $PCA(PosTgt(i))$
\EndFor

\For {each class $i$ \Pisymbol{psy}{206} $Object$ $Class$ }
\State $ProjectMat(i)$ $\gets$ $SubspaceAlign(X_{source}(i),X_{target}(i))$ 
\EndFor

\State $AdaptedRCNNdetector$ $\gets$ $TrainRCNNonSource(ProjectedSrcData)$
\State $boxes$ $\gets$ $runAdaptedRCNNdetector(ProjectedTgtData)$
\State $predictBbox$ $\gets$ $runNonMaximumSupression(boxes)$
\State \textbf{return} $predictBbox$
\EndProcedure

\end{algorithmic}
\end{algorithm}

Indeed the more specific subspace for each object category is expected to span the full space in which a particular object category lies, more accurately. Since we are dealing with the unsupervised setting hence we don't have access to bounding boxes of the target domain data. Source subspace can easily be found using the bounding boxes available for the source data. An important point to note here is that considering only the ground truth bounding boxes for getting the class specific subspace  in source domain data results in overfitting and the subspace obtained by this way doesn't truly represent the space of the specific class but represent a smaller and specific space which is subset of the original space. To avoid this problem, we consider all those bounding boxes whose overlap with the ground truth bounding box is above a particular threshold $\gamma$ while obtaining the class specific source domain subspaces. $\gamma$ is chosen by using cross validation on the source data. Once we obtain the class specific source subspaces, we also need to get the class specific target subspaces to apply subspace alignment method for domain adaptation. The problem here is that we don't have bounding boxes available for the target data and hence we direcly can not get the search windows and their overlap with the actual bounding boxes. To deal with this problem we run the RCNN-detector on target dataset which was initially trained on source dataset. Running the RCNN-detector gives the score for every search windows on the images of target dataset. We consider all the search windows of a specific class as positives samples for subspace generation whose score is greater than a certain threshold $\sigma$. Again this sigma is chosen by using cross validation on source dataset. Once we have positive samples for target subspace, we generate class specific target subspaces and apply subspace alignment approach on each class separately. The detector is trained separately for each class on target aligned source co-ordinate system. During the test, target data is projected on target subspace and classified using the detector trained on target aligned source co-ordinate system. In the last step, greedy non maximum suppression is applied to predict the most accurate window. Hence, the full algorithm is a two step process, as summarized in the algorithm \ref{alg:euclid}.

\vspace{-1.4cm}
\section{Experiments}
\label{expts}
\begin{figure}[H]
        \centering
        \begin{subfigure}[b]{0.15\textwidth}
                \includegraphics[width=2cm , height =2cm]{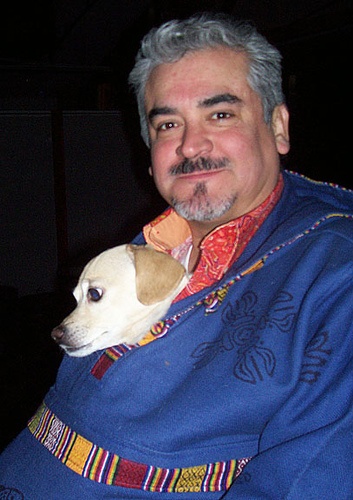}
                \caption{PASCAL}
        \end{subfigure}
        \begin{subfigure}[b]{0.15\textwidth}
                \includegraphics[width=2cm , height =2cm]{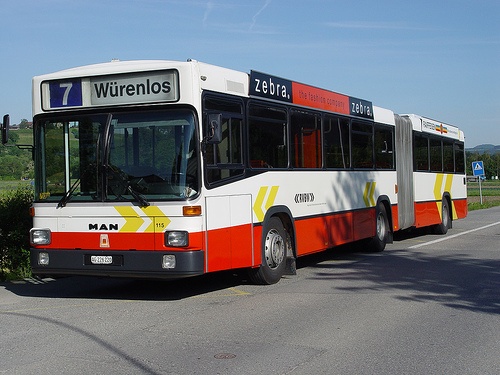}
                \caption{PASCAL}
        \end{subfigure} 
        \begin{subfigure}[b]{0.15\textwidth}
                \includegraphics[width=2cm , height =2cm]{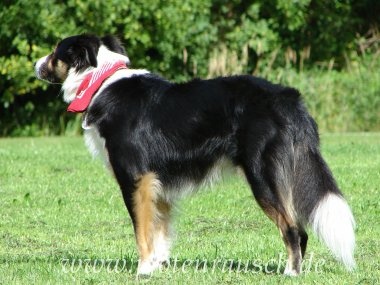}
                \caption{PASCAL}
        \end{subfigure}
        ~
        \begin{subfigure}[b]{0.15\textwidth}
                \includegraphics[width=2cm , height =2cm]{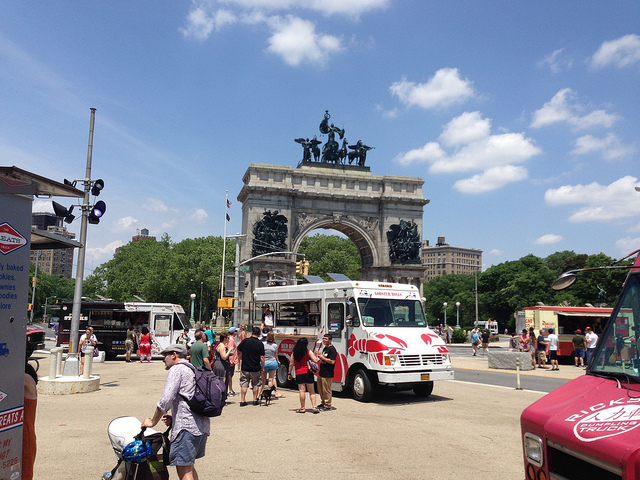}
                \caption{COCO}
        \end{subfigure}
        \begin{subfigure}[b]{0.15\textwidth}
                \includegraphics[width=2cm , height =2cm]{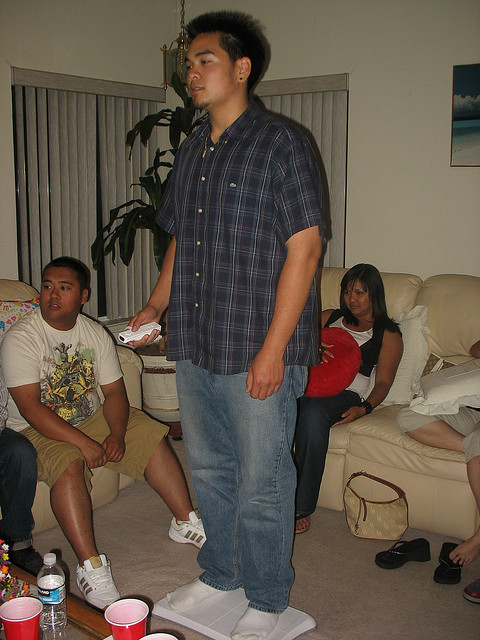}
                \caption{COCO}
        \end{subfigure}
        \begin{subfigure}[b]{0.15\textwidth}
                \includegraphics[width=2cm , height =2cm]{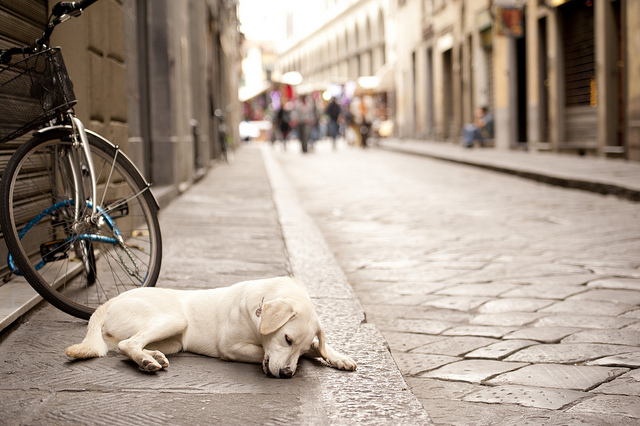}
                \caption{COCO }
        \end{subfigure}
        \vspace{0.1cm}
        \caption{Image samples taken from COCO validation set and PASCAL VOC 2012 to show domain shift}\label{fig:animals}
\end{figure}

\vspace{-0.2cm}
In this section we describe our experimental setup, dataset and then we evaluate the performance of our method. We analyze the performance of our added components and discuss its significance with the help of similarity in the principle angles of subspaces.  As Caffe features \cite{ES21} have recently resulted in state of the art for classification, we use Caffe features for evaluating our method. The initial set of positive examples for CoCo dataset are obtained by running R-CNN detector on the Validation set of CoCo. This R-CNN detector was trained on PASCAL VOC 2012. Here, Pascal VOC 2012 train+val dataset is the source and validation dataset of CoCo dataset is the target. The validation dataset of Microsoft CoCo dataset contains around 40,000 images and provides a challenging substantial dataset for comparison. While PASCAL dataset mostly consists of iconic view of the object, COCO dataset is more challenging and contains clutter, background, occlusion and multiple side views. These factors cause significant domain shift in the above mentioned two datasets and make these two dataset a suitable choice to evaluate our algorithm.  More details about the differences between PASCAL VOC 2012 and COCO dataset has been discussed in \cite{ES20} in great detail. We provide here in figure \ref{fig:animals}, some images from both the dataset to visualize the differences between them. It can be observed from the figure \ref{fig:animals} that while most of the images in PASCAL dataset contains single instance of object and less background clutter, images in COCO datset have generally more than one instance and also contain more background clutter.
\vspace{-0.1cm}

\vspace{-0.7cm}
\subsection{Experimental Setup}
PASCAL dataset is simpler than COCO and also smaller hence it has been considered as source dataset and COCO is considered as target dataset. For target dataset(COCO) subspace generation, the examples having a classifier score greater than a threshold of 0.4 are considered positive examples for the target subspace. Non maximum supression is removed from the process for consistency in the number of samples for subspace generation. For source dataset(PASCAL), we evaluate the overlap of each object proposal region with the ground truth bounding box and consider the bounding boxes with threshold greater than 0.7 with the ground truth bounding box as candidates for our source subspace generation. The dimensionality of subspace for both the source and target dataset is kept fixed at 100. Subspace alignment is done between source and target subspaces and source data set is projected on the target aligned source subspace for further training. Once the new detectors are trained on the transformed feature the same procedure is applied as RCNN for detection on projected target image features.
\vspace{-0.5cm}
\subsection{Results}

\begin{figure}[H]
        \centering
        \begin{subfigure}[b]{0.4\textwidth}
                {\includegraphics[width=6cm , height =3.8cm]{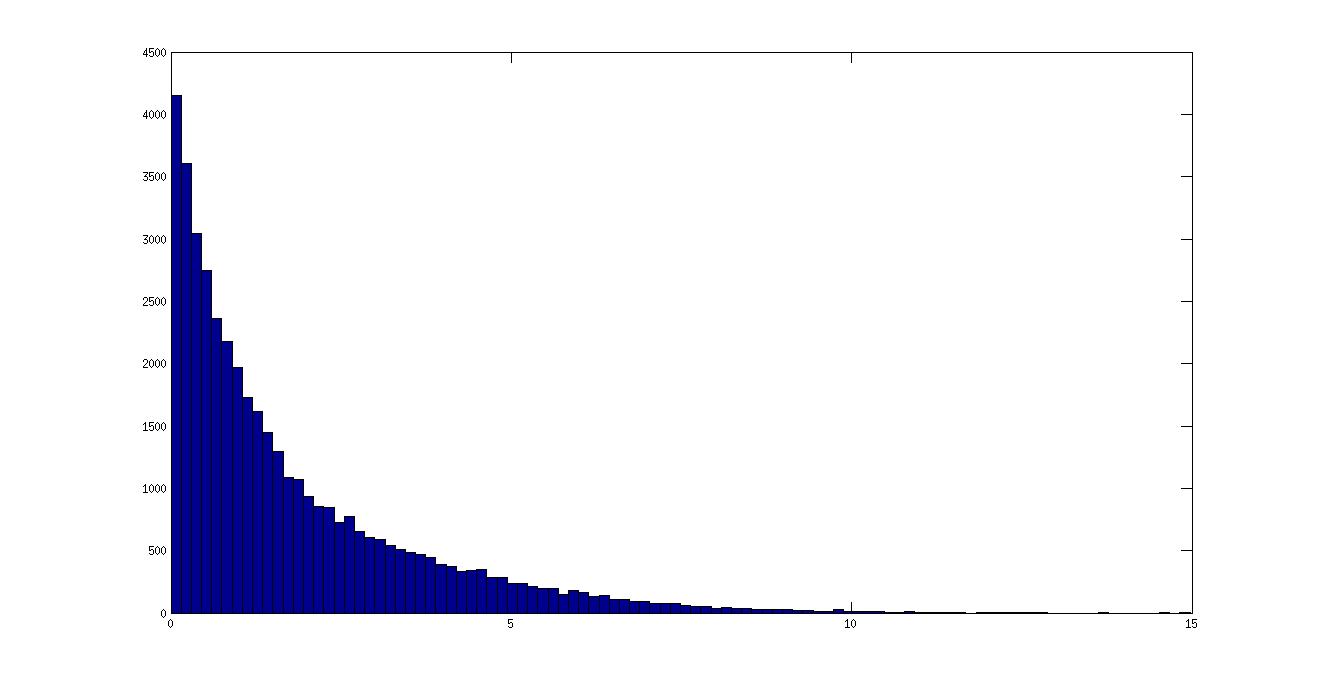}}

                \vspace{-0.2cm}
        \end{subfigure}
        ~
        ~
        \begin{subfigure}[b]{0.4\textwidth}
                {\includegraphics[width=6cm , height =3.8cm]{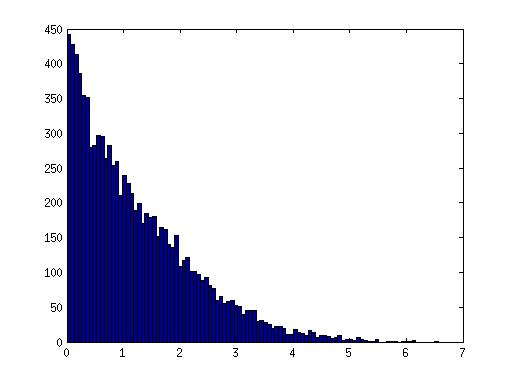}}

                \vspace{-0.2cm}
        \end{subfigure}
        \vspace{0.2cm}
        \caption{Histogram of scores. Fig 1 is for CoCo dataset and fig 2 for PASCAL VOC. Scores is taken along x-axis and no. of object region with that score along y-axis}
        \label{fig:histogram}
\end{figure}

\begin{figure}[H]
        \centering
        \begin{subfigure}[b]{0.15\textwidth}
                \includegraphics[width=2.5cm , height =2cm]{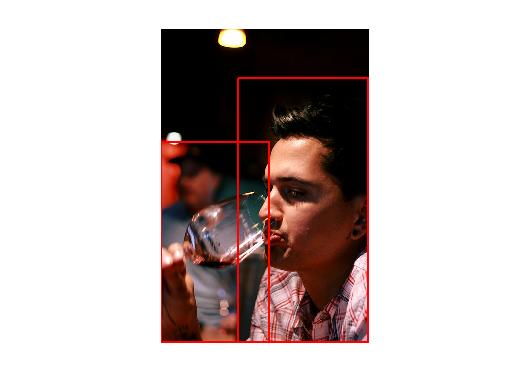}
             
        \end{subfigure}
        \begin{subfigure}[b]{0.15\textwidth}
                \includegraphics[width=2.5cm , height =2cm]{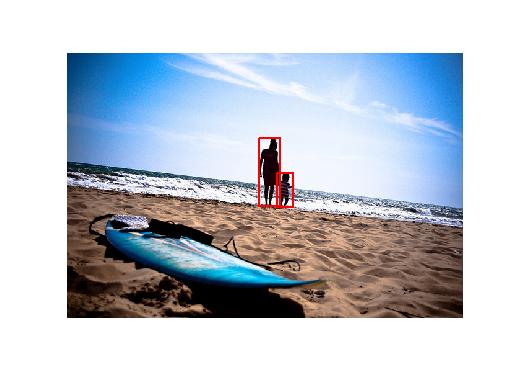}
                
        \end{subfigure} 
        \begin{subfigure}[b]{0.15\textwidth}
                \includegraphics[width=2.5cm , height =2cm]{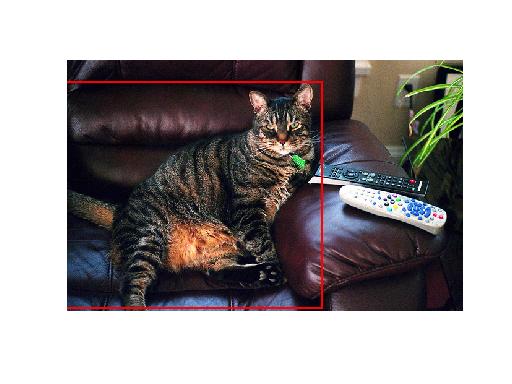}
                
        \end{subfigure}
        ~
        \begin{subfigure}[b]{0.15\textwidth}
                \includegraphics[width=2.5cm , height =2cm]{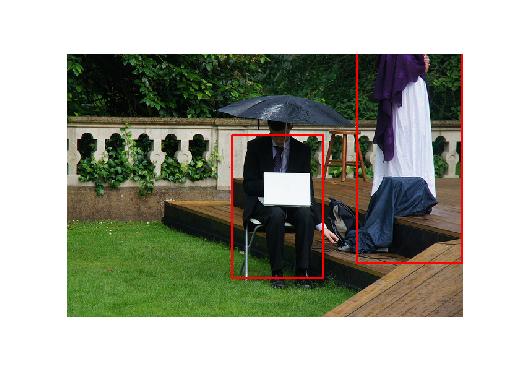}
                
        \end{subfigure}
        \begin{subfigure}[b]{0.15\textwidth}
                \includegraphics[width=2.5cm , height =2cm]{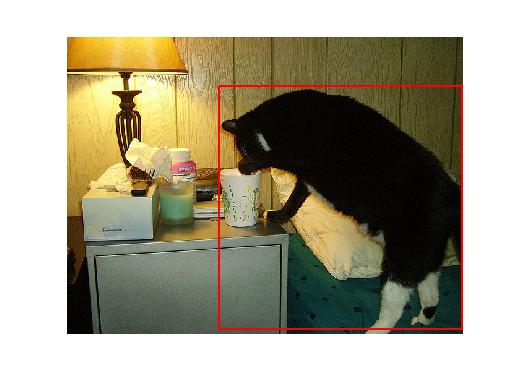}
                
        \end{subfigure}

        \vspace{0.1cm}
        \caption{Few extremely good detections using our proposed method}\label{good}
\end{figure}

\begin{figure}[H]
        \centering
        \begin{subfigure}[b]{0.15\textwidth}
                \includegraphics[width=2cm , height =2cm]{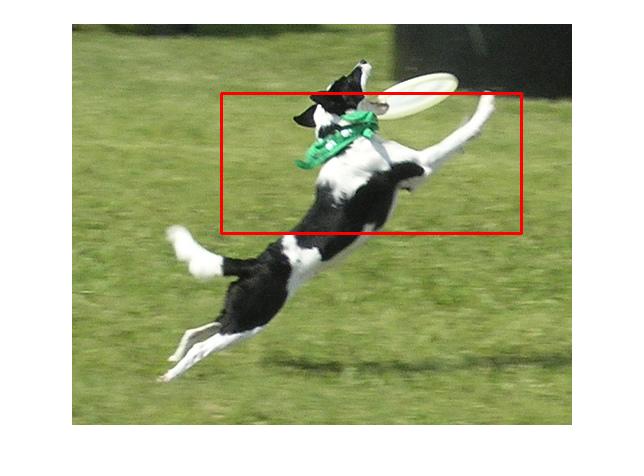}
                \caption{Our method}
        \end{subfigure}
        \begin{subfigure}[b]{0.15\textwidth}
                \includegraphics[width=2cm , height =2cm]{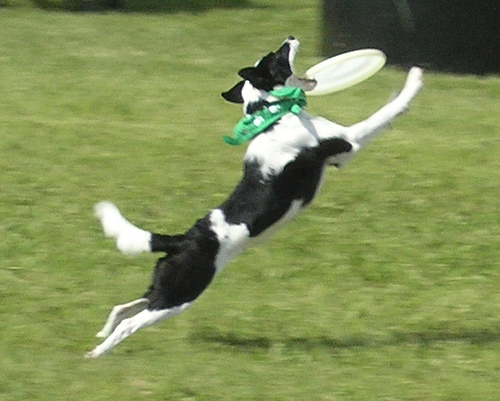}
                \caption{RCNN}
        \end{subfigure}         
        \begin{subfigure}[b]{0.15\textwidth}
                \includegraphics[width=2cm , height =2cm]{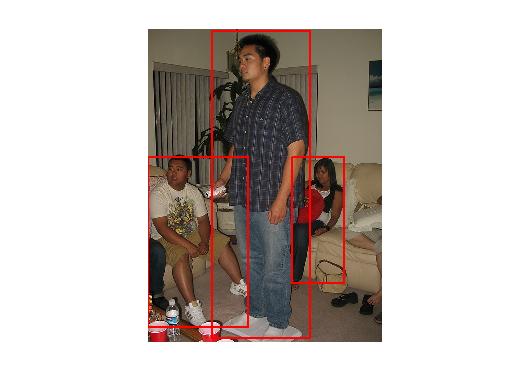}
                \caption{Our method}
        \end{subfigure}        
        \begin{subfigure}[b]{0.15\textwidth}
                \includegraphics[width=2cm , height =2cm]{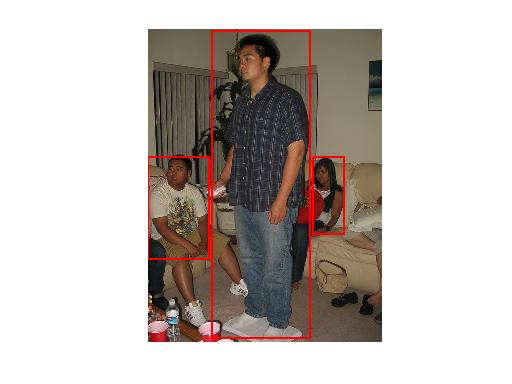}
                \caption{RCNN}
        \end{subfigure}
      
        \vspace{0.1cm}
        \caption{Examples where RCNN fails to perform but our method performs well}\label{compare}
\end{figure}

\begin{figure}[H]
        \centering
        \begin{subfigure}[b]{0.15\textwidth}
                \includegraphics[width=3cm , height =2cm]{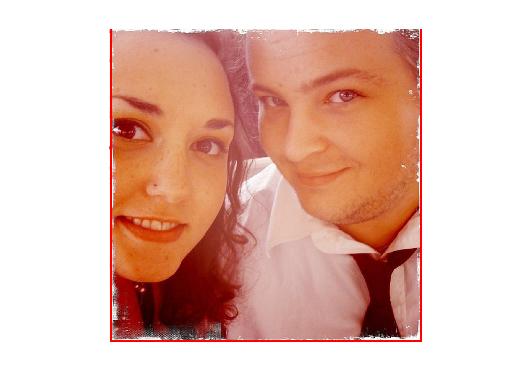}
                
        \end{subfigure}
        \begin{subfigure}[b]{0.15\textwidth}
                \includegraphics[width=3cm , height =2cm]{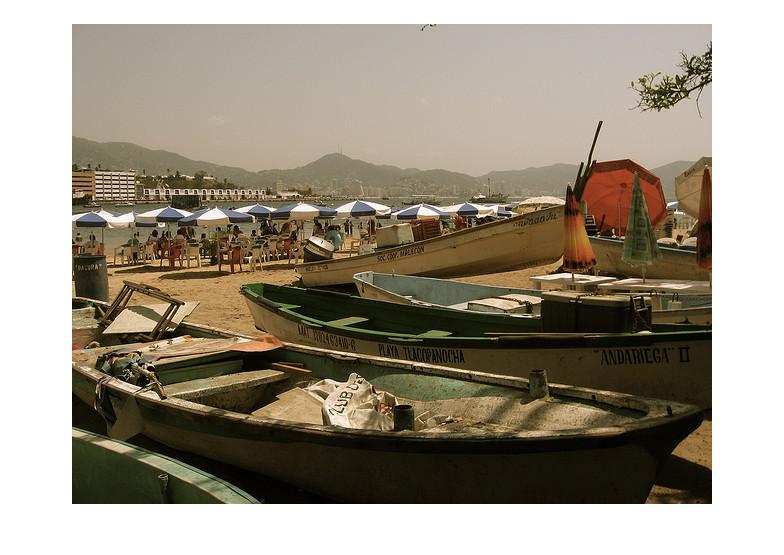}
                
        \end{subfigure} 
        \begin{subfigure}[b]{0.15\textwidth}
                \includegraphics[width=3cm , height =2cm]{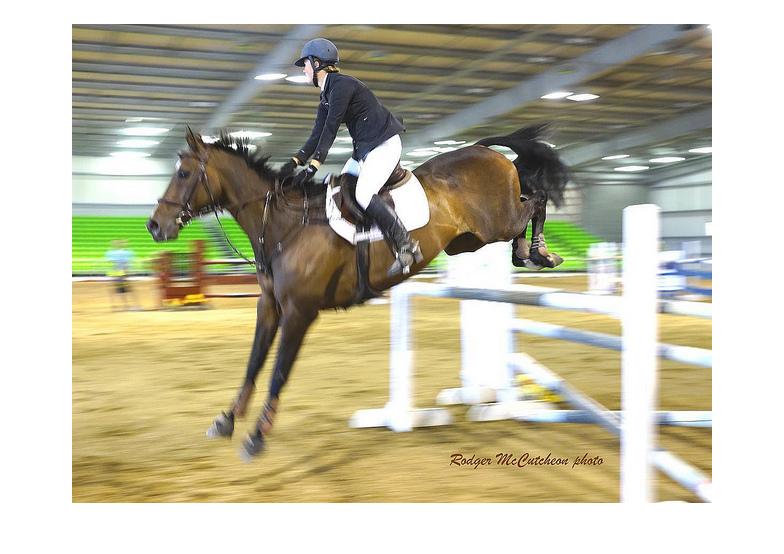}
                
        \end{subfigure}
        ~
        \begin{subfigure}[b]{0.15\textwidth}
                \includegraphics[width=3cm , height =2cm]{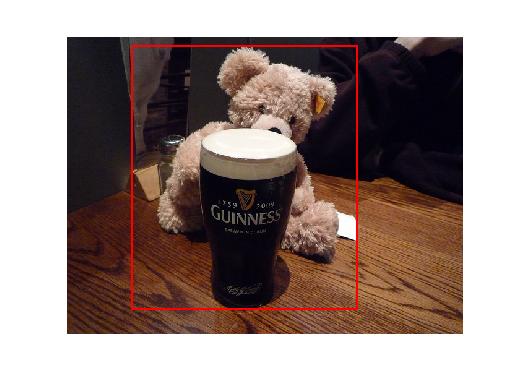}
                
        \end{subfigure}
        \vspace{0.1cm}
        \caption{Examples where both the detectors fails to perform. $4^{th}$ image is detected as human}\label{bad}
\end{figure}

Here in this section we provide evidence to show the  performance of our method, compare our results to other baselines and analyse the results. First we consider the statistical difference between the PASCAL VOC 2012 and COCO validation set data. We run RCNN detector on both source and target data. We plot the histograms of score obtained for both the dataset. It can be observed from their histogram in image \ref{fig:histogram}  that there exists statistical dissimilarity between both these datasets. Therefore, there is a need for domain adaptation. The histogram evaluation has been done in two setting, first in a category wise setting and second as a full dataset jointly. The findings of both these settings is similar and demonstrates the statistical dissimilarity between these two datasets. 
\vspace{-0.35cm}
\vspace{-0.3cm}
Now we evaluate our method on these two datasets. First baseline to compare with our method is a simple RCNN detector trained on PASCAL VOC 2012. We evaluate the performance of RCNN detector for all the 20 categories which are there in the PASCAL datasets. The mean average precision for this baseline comes to 23.60\% without applying bounding box regression. Second baseline to compare with our method is RCNN - full image transform. RCNN - full image transform here indicates that while aligning the subspaces for source and target data, we don't consider class specific subspace alignment. We take the full target and source data at once, learn separate subspaces for source and target data, apply the transformation on source domain subspace to align with the target domain subspace and in the last step we train the detector with projected features in target aligned source co-ordinate system. This baseline is similar to the work done for unsupervised domain adaptation of object classifier using subspace alignment \cite{ES3} as discussed in section \ref{SA}. The mean average precision using RCNN - full image transform is 22.70\%.  We also compare our result with the result obtained on the COCO validation set using deformable part model trained on PASCAL VOC 2012 \cite{ES22} as reported in \cite{ES20}. The mean average precision for deformable part model is reported as 16.9\%. We got consistent improvement over all the baselines using our proposed approach. The proposed approach, RCNN - local class specific transform, gives the mean average precision of \textbf{25.43\%} which is almost 1.8\% better than the traditional RCNN detector and 8.5\% higher than the deformable part model based object detector. The complete result with category wise performance is given in table \ref{the_table}. We also show here some of the accurate detections obtained using our method and visually compare the detection by traditional RCNN object detector on the same images. In figure \ref{good} we show some detections obtained using our proposed method. Figure \ref{compare} is used to illustrate the cases when our proposed method works but RCNN fails. Figure \ref{bad} contains some images both the detectors failed to perform. In the next subsection \ref{PAA}, we discuss the intuition behind our performance and explain it  with the principal angle analysis of source and target subspaces.
\begin{table}[htb]
\centering
\begin{tabular}{|l|l|c|c|c|c|}
\hline
No. & class & RCNN-  & RCNN -  & Proposed, RCNN - &DPMv5-P\\
& &No Transform & Full Transform & Local Transform & \\
\hline
1& plane&              36.72    & 35.44      & \textbf{40.1} & 35.1 \\
2&bicycle&             21.26      &18.95       &  \textbf{23.28} &1.9             \\
3&bird&                 12.50      &12.37       & \textbf{13.63}  &3.7            \\
4&boat&                 10.45       &8.8         & \textbf{10.61}  &2.3     \\
5&bottle&              8.75        &\textbf{11.46} &   8.11        &7     \\
6&bus&                 37.47       &38.12       &   40.64 &\textbf{45.4}           \\
7&car&                 20.6        & 20.4       & \textbf{22.5} &18.3                \\
8&cat&                 42.4        &  43.6      &  \textbf{45.6}  &8.6              \\
9&chair&              \textbf{9.6}          &  6.3       &  8.8  &6.3              \\
10&cow&               23.28        &  20.40     &     \textbf{25.3}  &17           \\
11&table&             15.9         &  14.9       &   \textbf{17.3}   &4.8            \\
12&dog&               28.42        &  \textbf{32.72}     &    31.3  &5.8           \\
13& horse &           30.7         &  31.11     &    32.9 &\textbf{35.3}            \\
14&motorbike &        31.2         &  29.05     &    \textbf{34.6}  &25.4           \\
15&person&            27.8         &    28.8        &  \textbf{30.9}  &17.5             \\
16&plant&             12.65        &  7.34      &   \textbf{13.7} &4.1             \\
17&sheep&            19.99         &  21.04     &  \textbf{22.4}  &14.5             \\
18&sofa&             14.6          &  8.4       & \textbf{15.5}  &9.6              \\
19&train&            39.2          & 38.4       & \textbf{41.64}   &31.7            \\
20&tv&               28.6          & 26.4       & \textbf{29.9}   &27.9             \\
\hline
 & Mean AP          &23.60         &   22.7         &\textbf{25.43}      &16.9      \\
\hline
\end{tabular}
\vspace{0.2cm}
\caption{Domain adaption detection result on validation set of COCO dataset. RCNN- No Transform column represent running the simple RCNN detector on COCO dataset which has been trained on PASCAL VOC 2012. RCNN- Full Transform denotes the result obtained by retraining the detector while considering the full source images and target images for aligning the subspace of source and target dataset. RCNN- Local Transform means the category specific subspace alignment method (proposed method) for adapting the detector and DPM5-P denotes the result from deformable part model on COCO dataset, trained on PASCAL VOC 2012 as reported in \cite{ES20}.}
\label{the_table}
\end{table}

\vspace{-0.8cm}

\vspace{-0.8cm}

\vspace{-0.8cm}
\subsubsection{Principal Angle Analysis}
\label{PAA}

As is evident from the table \ref{the_table}, the proposed method outperforms the baselines for almost all of the classes except for a few categories. We also analyse category wise similarity between the source subspaces and target subspaces to explain this phenomena. The similarity between 
\begin{wrapfigure}{r}{0.6\textwidth}
\vspace{-35pt}
  \begin{center}
    {\includegraphics[width=7.5cm,height=7cm]{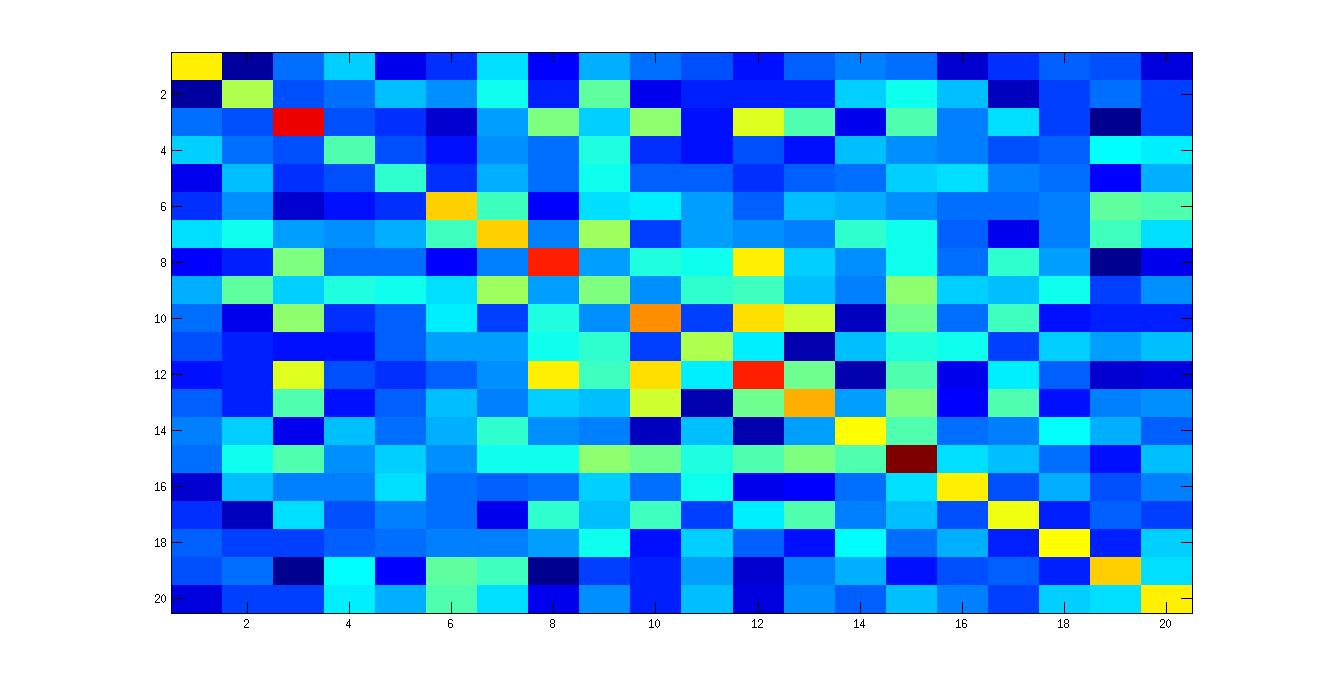}}
  \end{center}
  \vspace{-20pt}
  \caption{Color map of similarity between learned subspaces of different categories in source and target dataset}
   \vspace{-50pt}
  \label{colormap}
\end{wrapfigure}
two subspaces is calculated by first finding the principal angles between two subspaces and then taking the 2-norm of cosines of vectors of principal angles between them.

\begin{center}
$d(X_s,X_t)$ = $\|Cos\Theta\|_2$, where $\Theta$ is the vector of principal angles between subspaces $X_s$ and $X_t$
\end{center}

In the results we observe that our method is either improving very little or not improving at all for which the traditional RCNN detector itself is not performing well. Our method relies on the initial detection obtained by the RCNN. When the initial detection is better, then, we learn a better subspace and hence we obtain a better subspace alignment. For classes where initial detections are not good then no meaningful subspace can be learned for such categories. This can be observed for class no. $4$(boat), $5$(bottle) and $9$(chair) from figure \ref{colormap} that similarity between source subspace of those classes and target subspace of the corresponding classes are very low. Hence, the performance of our method on these classes are not good. But for the rest of the classes, as can be seen from the diagonal blocks of the figure \ref{colormap}, the target subspaces are quite discriminative and inter class subspaces are not very similar. This gives improvement in the result. One more interesting thing to discuss here is that for class no. $12$, though the similarity between source subspace for class no. $12$ and target subspace for same class is good but this subspace is also very similar to the subspaces of other categories as a few numbers of yellow blocks are available there in the row and column corresponding to class $12$. In that  region, full image transformation is expected to perform better for this class and our experimental results also demonstrate this point.   


\section{Conclusion}
\label{concl}
In this paper, we have presented a method for adapting the state-of-the-art RCNN object detector for unsupervised domain adaptation from source to target dataset. The main challenge addressed in this paper is to obtain localized domain adaptation for adapting object detectors. The adaptation has been achieved by using approach based on subspace alignment that efficiently projects the source subspace to the target subspace. The proposed method results in improved object detection. Thus, one can use RCNN for instance to detect persons in novel settings with improved detection accuracy. 

The main limitation of this approach is that the present method does not work well for classes where the RCNN results are weak. This limitation can be addressed by partially relying on supervision for classes where the source detection result is itself quite weak. Further, once there is some supervision available, it would be interesting to jointly consider learning domain adaptation at both feature and subspace level simultaneously.

\section{Acknowledgements}
The authors would like to acknowledge funding support received from the ERC starting grant Cognimund and the Research-I foundation at the Computer Science department at IIT Kanpur

\end{document}